\documentclass[11pt,a4paper]{article}
\usepackage{authblk}
\usepackage[hyperref]{acl2021}
\usepackage{times}
\usepackage{latexsym}
\usepackage{graphicx}
\usepackage{subcaption}
\usepackage{verbatim}
\usepackage{multirow}
\usepackage{svg}
\usepackage[normalem]{ulem}

\graphicspath{ {./img/} }

\usepackage{microtype}

\aclfinalcopy 
\def\aclpaperid{631} 

\title{Investigating Text Simplification Evaluation}

\author{\textbf{Laura Vásquez-Rodríguez}\textsuperscript{\textnormal{1}}, 
        \textbf{Matthew Shardlow}\textsuperscript{\textnormal{2}},
        \textbf{Piotr Przybyła}\textsuperscript{\textnormal{3}},
        \textbf{Sophia Ananiadou}\textsuperscript{\textnormal{1}}\vspace{-3mm}\\
  \textsuperscript{\textnormal{1}}National Centre for Text Mining,\\The University of Manchester, Manchester, United Kingdom \\
  \textsuperscript{\textnormal{2}}Department of Computing and Mathematics, \\Manchester Metropolitan University, Manchester, United Kingdom  \\
  \textsuperscript{\textnormal{3}}Institute of Computer Science, Polish Academy of Sciences, Warsaw, Poland \\
  {\tt \{laura.vasquezrodriguez, sophia.ananiadou\}@manchester.ac.uk} \\
  {\tt m.shardlow@mmu.ac.uk} \hspace{1em}
  {\tt piotr.przybyla@ipipan.waw.pl} \\
 }

\date{}

\begin{document}
\maketitle
\begin{abstract}
Modern text simplification (TS) heavily relies on the availability of gold standard data to build machine learning models. However, existing studies show that parallel TS corpora contain inaccurate simplifications and incorrect alignments. Additionally, evaluation is usually performed by using metrics such as BLEU or SARI to compare system output to the gold standard. A major limitation is that these metrics do not match human judgements and the performance on different datasets and linguistic phenomena vary greatly. 
Furthermore, our research shows that the test and training subsets of parallel datasets differ significantly. In this work, we investigate existing TS corpora, providing new insights that will motivate the improvement of existing state-of-the-art TS evaluation methods. Our contributions include the analysis of TS corpora based on existing modifications used for simplification and an empirical study on TS models performance by using better-distributed datasets. We demonstrate that by improving the distribution of TS datasets, we can build more robust TS models.

\end{abstract}

\section{Introduction}
\label{introduction}

Text Simplification transforms natural language from a complex to a simple format, with the aim to not only reach wider audiences~\citep{Rello2013, DeBelder2010, Aluisio2010, Inui2003} but also as a preprocessing step in related tasks~\citep{Shardlow2014, Silveira2012}.

Simplifications are achieved by using parallel datasets to train sequence-to-sequence text generation algorithms~\citep{Nisioi2017} to make complex sentences easier to understand. 
They are typically produced by crowdsourcing~\citep{Xu2016,Alva-Manchego2020b} or by alignment \cite{Cao2020,Jiang2020}. They are infamously noisy and models trained on these give poor results when evaluated by humans~\citep{cooper-shardlow-2020-combinmt}. 
In this paper we add to the growing narrative around the evaluation of natural language generation \cite{van-der-lee-etal-2019-best,caglayan-etal-2020-curious,Pang2019}, focusing on parallel text simplification datasets and how they can be improved.

\textbf{Why do we need to re-evaluate TS resources?}

In the last decade, TS research has relied on Wikipedia-based datasets~\citep{Zhang2017,Xu2016,Jiang2020}, despite their known limitations \cite{Xu2015, Alva-Manchego2020b} such as questionable sentence pairs alignments, inaccurate simplifications and a limited variety of simplification modifications.
Apart from affecting the reliability of models trained on these datasets, their low quality influences the evaluation relying on automatic metrics that requires gold-standard simplifications, such as SARI~\citep{Xu2016} and BLEU~\citep{Papineni2002}. 

Hence, evaluation data resources must be further explored and improved to achieve reliable evaluation scenarios. There is a growing body of evidence~\citep{Xu2015} (including this work) to show that existing datasets do not contain accurate and well-constructed simplifications, significantly impeding the progress of the TS field. 

Furthermore, well-known evaluation metrics such as BLEU are not suitable for simplification evaluation. According to previous research~\citep{Sulem2018} BLEU does not significantly correlate with simplicity~\citep{Xu2016}, making it inappropriate for TS evaluation. Moreover, it does not correlate (or the correlation is low) with grammaticality and meaning preservation when performing syntactic simplification such as sentence splitting. Therefore in most recent TS research BLEU has not been considered as a reliable evaluation metric. We use SARI as the preferred method for TS evaluation, which has also been used as the standard evaluation metric in all the  corpora analysed in this research.

Our contributions include 1) the analysis of the most common TS corpora based on quantifying modifications used for simplification, evidencing their limitations and 2) an empirical study on TS models performance by using better-distributed datasets. We demonstrate that by improving the distribution of TS datasets, we can build TS models that gain a higher SARI score in our evaluation setting.

\section{Related Work}

The exploration of neural networks in TS started with the work of \citet{Nisioi2017}, using the largest parallel simplification resource available \cite{Hwang2015}.  Neural-based work focused on state-of-the-art deep learning and MT-based methods, such as reinforcement learning~\citep{Zhang2017}, adversarial training~\citep{Surya2019}, pointer-copy mechanism~\citep{Guo2018}, neural semantic encoders~\citep{Vu2018} and transformers supported by paraphrasing rules~\citep{Zhao2018}. 

Other successful approaches include the usage of control tokens to tune the level of simplification expected \citep{Alva-Manchego2020b, Scarton2018} and the prediction of operations using parallel corpora \citep{Alva-Manchego2017, Dong2020}. The neural methods are trained mostly on Wikipedia-based sets, varying in size and improvements in the quality of the alignments.  

\citet{Xu2015} carried out a systematic study on Wikipedia-based simplification resources, claiming Wikipedia is not a quality resource, based on the observed alignments and the type of simplifications. \citet{Alva-Manchego2020b} proposed a new dataset, performing a detailed analysis including edit distance and proportion of words that are deleted, inserted and reordered, and evaluation metrics performance for their proposed corpus. 

Chasing the state-of-the-art is rife in NLP \cite{hou-etal-2019-identification}, and no less so in TS, where a SARI score is too often considered the main quality indicator. 
However, recent work has shown that these metrics are unreliable \cite{caglayan-etal-2020-curious} and gains in performance according to them may not deliver improvements in simplification performance when the text is presented to an end user. 

\begin{figure*}
\centering
\begin{subfigure}[b]{.30\linewidth}
\includegraphics[width=\linewidth]{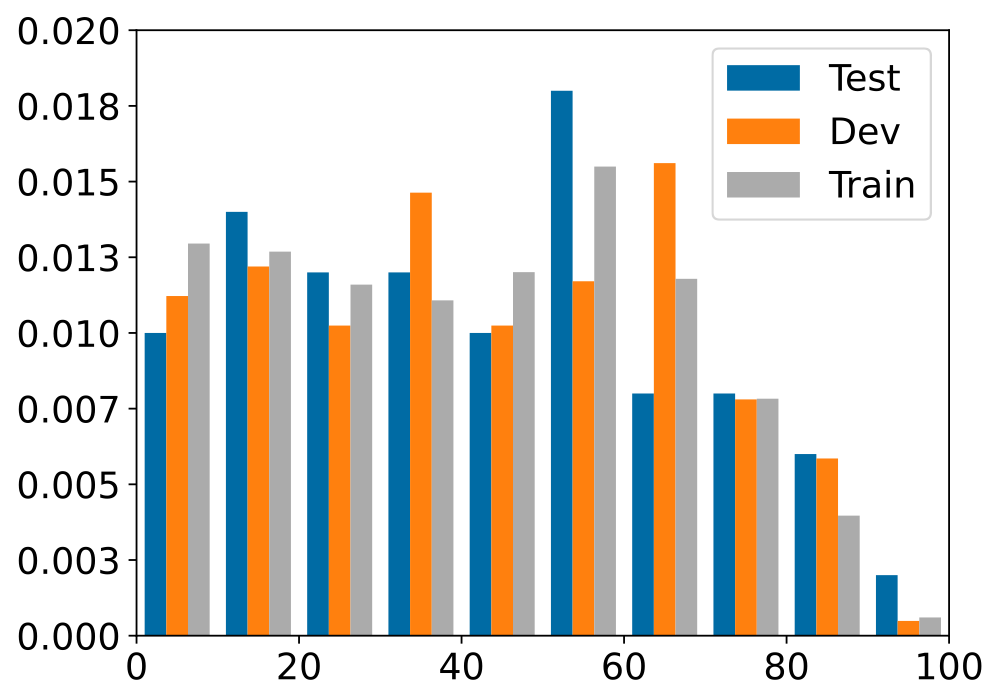}
\caption{WikiSmall Test/Dev/Train}\label{fig:wikismall_ed}
\end{subfigure}
\begin{subfigure}[b]{.30\linewidth}
\includegraphics[width=\linewidth]{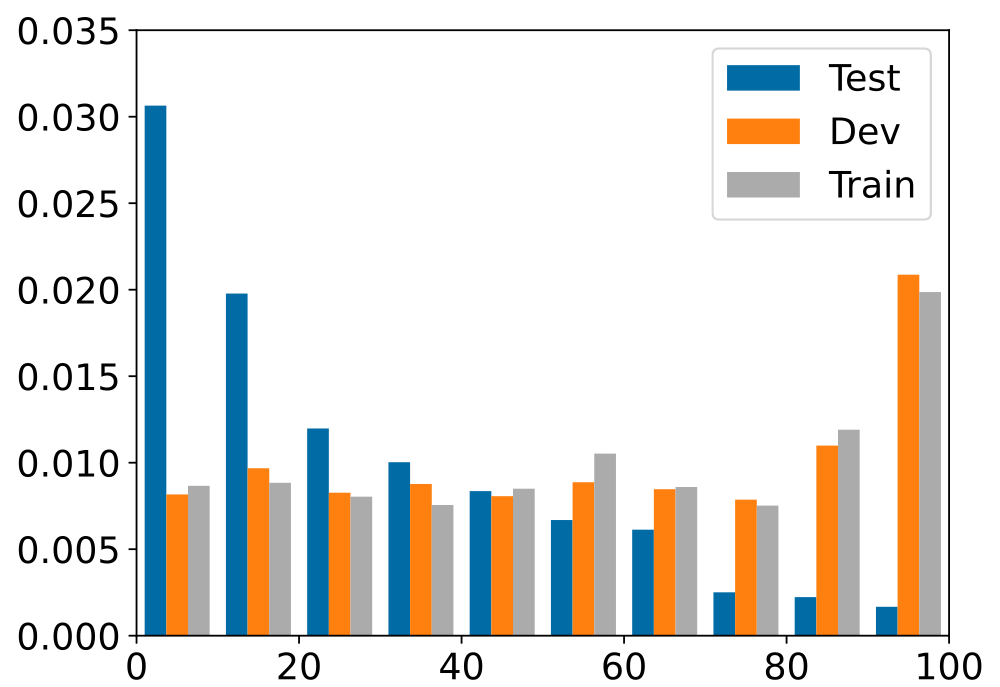}
\caption{WikiLarge Test/Dev/Train}\label{fig:wikilarge_ed}
\end{subfigure}
\begin{subfigure}[b]{.30\linewidth}
\includegraphics[width=\linewidth]{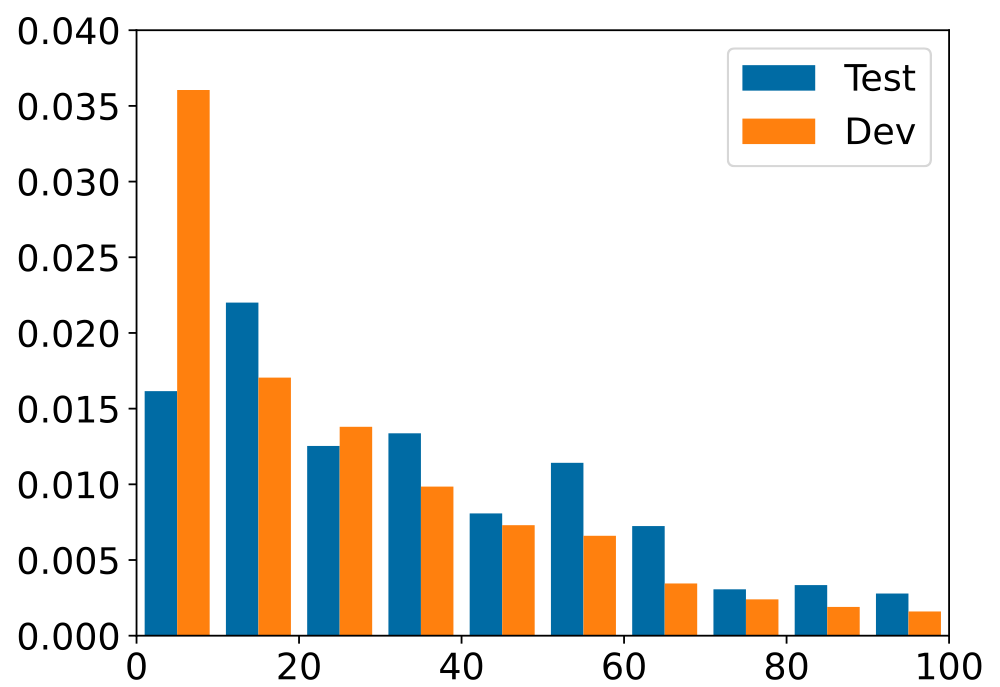}
\caption{TurkCorpus Test}\label{fig:turkcorpus_ed}
\end{subfigure}

\begin{subfigure}[b]{.30\linewidth}
\includegraphics[width=\linewidth]{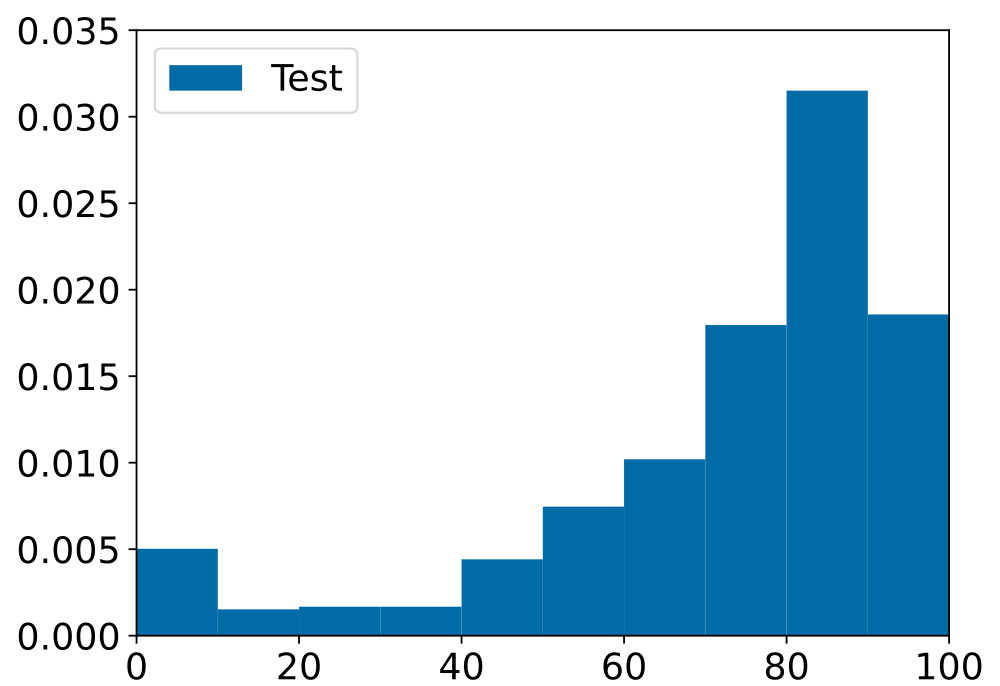}
\caption{MSD Test}\label{fig:msd_ed}
\end{subfigure}
\begin{subfigure}[b]{.30\linewidth}
\includegraphics[width=\linewidth]{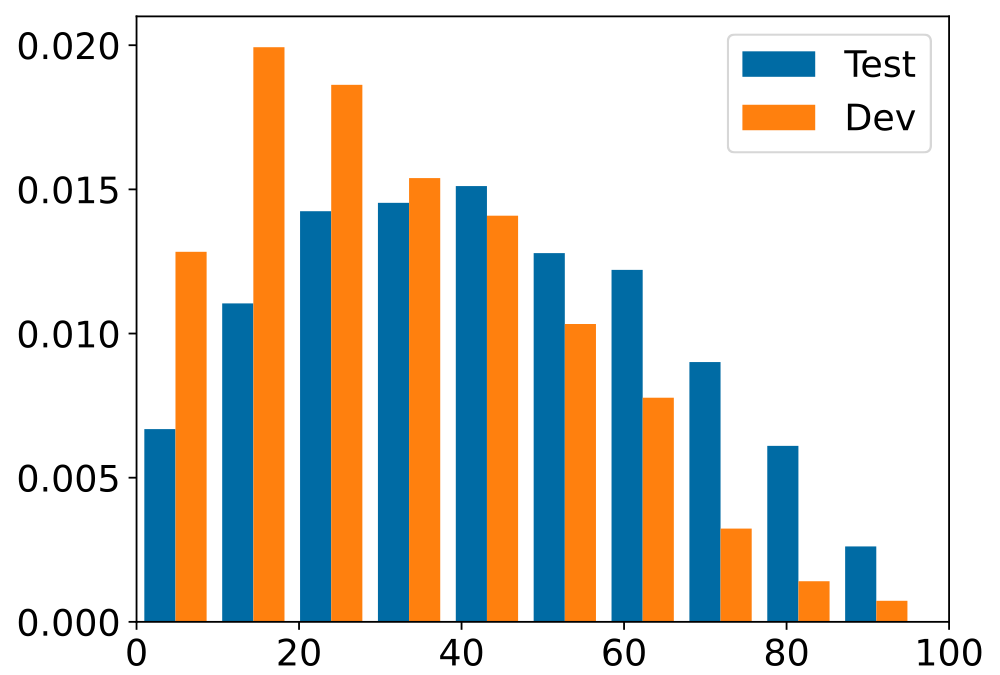}
\caption{ASSET Test}\label{fig:asset_ed}
\end{subfigure}
\begin{subfigure}[b]{.30\linewidth}
\includegraphics[width=\linewidth]{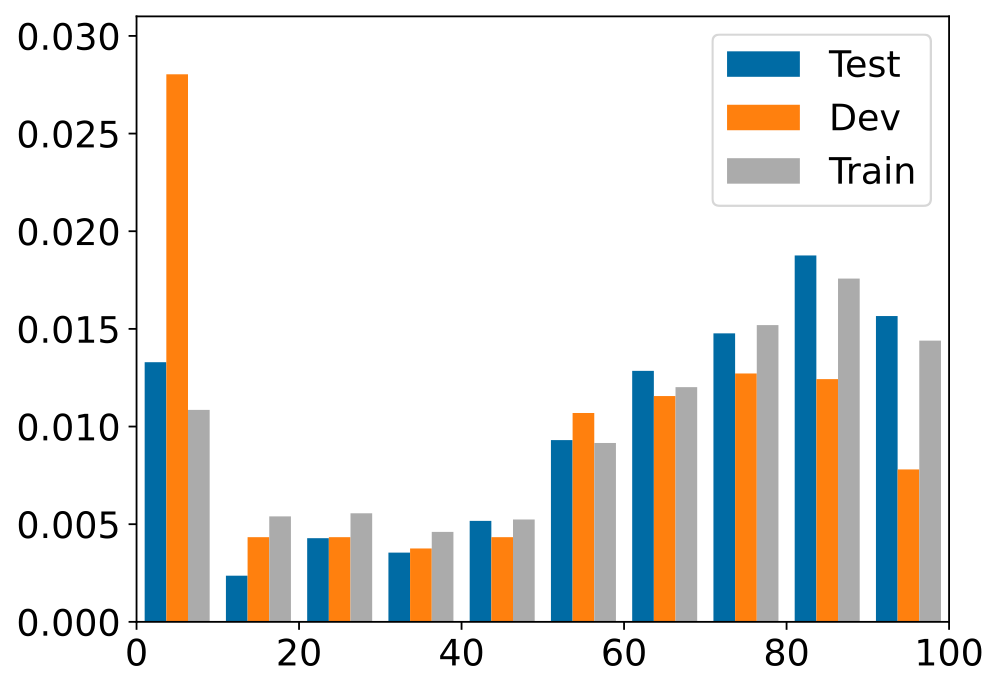}
\caption{WikiManual Test/Dev/Train}\label{fig:wikimanual_ed}
\end{subfigure}
\caption{Comparison of TS datasets with respect to the number of edit operations between the original and simplified sentences. X-axis: token edit distance normalised by sentence length, Y-axis: probability density for the change percentage between complex and simple sentence pairs.}
\label{fig:edit_distance}
\end{figure*}

\section{Simplification Datasets: Exploration}

\subsection{Data and Methods}

In the initial exploration of TS datasets, we investigated the training, test and validation subsets (when available) of the following: WikiSmall and WikiLarge~\citep{Zhang2017}, TurkCorpus~\citep{Xu2015}, MSD dataset~\citep{Cao2020}, ASSET~\citep{Alva-Manchego2020b} and WikiManual~\citep{Jiang2020}. For the WikiManual dataset, we only considered sentences labelled as ``aligned".

We computed the number of changes between the original and simplified sentences through the \textbf{token edit distance}. Traditionally, edit distance quantifies character-level changes from one character string to another (additions, deletions and replacements). In this work, we calculated the token-based edit distance by adapting the Wagner–Fischer algorithm~\citep{Wagner1974} to determine changes at a token level. We preprocessed our sentences by changing them into lowercase prior to this analysis. To make the results comparable across sentences, we divide the number of changes by the length of the original sentence and obtain values between 0\% (no changes) to 100\% (completely different sentence).

In addition to toked-based edit operation experiments, we analysed the difference of sentence length between complex and simple variants, the quantity of edit operations type (INSERT, DELETE and REPLACE) and an analysis of redundant operations such as deletions and insertions in the same sentence over the same text piece (we define this as the MOVE operation). Based on our objective to show how different split configurations affect TS model performance, we have presented the percentage of edit operations as the more informative analysis performed on the most representative datasets. 

\subsection{Edit Distance Distribution}
\label{sec:ts_datasets}

Except for the recent work of \citet{alva-manchego-etal-2020-asset}, there has been little work on new TS datasets.
Most prior datasets are derived by aligning English and Simple English Wikipedia, for example \textbf{WikiSmall} and \textbf{WikiLarge}~\citep{Zhang2017}.  

In Figure~\ref{fig:edit_distance} we can see that the edit distance distribution of the splits in the selected datasets is not even. By comparing the test and development subsets in WikiSmall (Figure~\ref{fig:wikismall_ed}) we can see differences in the number of modifications involved in simplification. Moreover, the WikiLarge dataset (Figure~\ref{fig:wikilarge_ed}) shows a complete divergence of the test subset. Additionally, it is possible to notice a significant number of unaligned or noisy cases, between the 80\% and 100\% of change in the WikiLarge training and validation subsets (Figure~\ref{fig:wikilarge_ed}). 

We manually checked a sample of these cases and confirmed they were poor-quality simplifications, including incorrect alignments. The simplification outputs (complex/simple pairs) were sorted by their edit distances and then manually checked to determine an approximate heuristic for noisy sentences detection. Since many of these alignments had really poor quality, it was easy to determine the number that removed a significant number of cases without actually reducing dramatically the size of the dataset.

Datasets such as \textbf{Turk Corpus}~\citep{Xu2015} are widely used for evaluation and their operations mostly consist of lexical simplification~\citep{Alva-Manchego2020b}. We can see this behaviour in Figure~\ref{fig:turkcorpus_ed}, where most edits involve a small percentage of the tokens. This can be noticed when a large proportion of the sample cases are between 0\% (no change) to 40\%. 

In the search of better evaluation resources, TurkCorpus was improved with the development of \textbf{ASSET}~\citep{Alva-Manchego2020b} including more heterogeneous modification measures. As we can see in Figure~\ref{fig:asset_ed}, the data are more evenly distributed than in Figure~\ref{fig:turkcorpus_ed}.

Recently proposed datasets, such as \textbf{WikiManual}~\citep{Jiang2020}, as shown in Figure~\ref{fig:wikimanual_ed}, have an approximately consistent distribution, and their simplifications are less conservative. Based on a visual inspection on the uppermost values of the distribution ($\approx$80\%), we can tell that often most of the information in the original sentence is removed or the target simplification does not express accurately the original meaning.

\textbf{MSD dataset}~\citep{Cao2020} is a domain-specific dataset, developed for style transfer in the health domain. 
In the style transfer setting, the simplifications are aggressive (i.e., not limited to individual words), to promote the detection of a difference between one style (expert language) and another (lay language). Figure~\ref{fig:msd_ed} shows how their change-percentage distribution differs dramatically in comparison to the other datasets, placing most of the results at the right-side of the distribution.

Among TS datasets, it is important to mention that the raw text of the \textbf{Newsela}~\citep{Xu2015} dataset was produced by professional writers and is likely of higher quality than other TS datasets. Unfortunately, it is not aligned at the sentence level by default and its usage and distribution are limited by a restrictive data agreement. We have not included this dataset in our analysis due to the restrictive licence under which it is distributed.  

\subsection{KL Divergence}
\label{sec:kl_divergence_results}

\begin{table}
 \centering
 \begin{tabular}{c|c|c|c} 
 \hline
 \hline
 \textbf{Dataset} & \textbf{Split} & \textbf{KL-div} & \textbf{p-value} \\
 \hline
 \multirow{2}{*}{WikiSmall} 
 & Test/Dev & 0.0696 & 0.51292 \\
 & Test/Tr & 0.0580 & 0.83186 \\
 \hline
 \multirow{2}{*}{WikiLarge} 
 & Test/Dev & 0.4623 & {\textless0.00001} \\
 & Test/Tr & 0.4639 & {\textless0.00001} \\
 \hline
 \multirow{2}{*}{WikiManual} 
 & Test/Dev & 0.1020 & 0.00003 \\
 & Test/Tr & 0.0176 & 0.04184 \\  
 \hline
 TurkCorpus 
 & Test/Dev & 0.0071 & 0.00026 \\
 \hline
 ASSET 
 & Test/Dev & 0.0491 & {\textless0.00001} \\
 \hline
 \hline
 \end{tabular}
 \caption{KL-divergence between testing (Test) and development (Dev) or training (Tr) subsets.}
 \label{kl-table}
 \vspace{-4mm} 
 \end{table}
 
In addition to edit distance measurements presented in Figure~\ref{fig:edit_distance}, we further analysed \textbf{KL divergence} \citep{Kullback1951} of those distributions to understand how much dataset subsets diverge.

Specifically, we compared the distribution of the test set to the development and training sets for WikiSmall, WikiLarge, WikiManual, TurkCorpus and ASSET Corpus (when available). We did not include MSD dataset since it only has a testing set.

We performed \textbf{randomised permutation tests}~\citep{Morgan2006} to confirm the statistical significance of our results. Each dataset was joined together and split randomly for 100,000 iterations. We then computed the $p$-value as a percentage of random splits that result in the KL value equal to or higher than the one observed in the data. 
Based on the $p$-value, we can decide whether the null hypothesis (i.e. that the original splits are truly random) can be accepted. We reject the hypothesis for $p$-value lower than 0.05.
In Table \ref{kl-table} we show the computed KL-divergence and $p$-values. The $p$-values below 0.05 for WikiManual and WikiLarge  confirm that these datasets do not follow a truly random distribution.

\section{Simplification Datasets: Experiments}
We carried out the following experiments to evaluate the variability in performance of TS models caused by the issues described in Wiki-based data.

\subsection{Data and Methods}

For the proposed experiments, we used the EditNTS model, a Programmer-Interpreter Model~\citep{Dong2020}. Although the original code was published, its implementation required minor modifications to run in our setting. The modifications performed, the experimental subsets as well as the source code are documented via GitHub\footnote{\url{https://github.com/lmvasque/ts-explore}}. We selected EditNTS model due to its competitive performance in both WikiSmall and WikiLarge datasets\footnote{\url{https://github.com/sebastianruder/NLP-progress/blob/master/english/simplification.md}}. Hence, we consider this model as a suitable candidate for evaluating the different limitations of TS datasets. In future work, we will definitely consider testing our assumptions under additional metrics and models. 

In relation to TS datasets, we trained our models on the training and development subsets from WikiLarge and WikiSmall, widely used in most of TS research. In addition, these datasets have a train, development and test set, which is essential for retraining and testing the model with new split configurations. The model was first trained with the original splits, and then with the following variations:

\textbf{Randomised split}: as explained in Section \ref{sec:kl_divergence_results}, the original WikiLarge split does not have an even distribution of edit-distance pairs between subsets. For this experiment, we resampled two of our datasets (WikiSmall and WikiLarge). For each dataset, we joined all subsets together and performed a new random split.  

\textbf{Refined and randomised split}: we created subsets that minimise the impact of poor alignments. These alignments were selected by edit distance and then subsets were randomised as above. We presume that the high-distance cases correspond to noisy and misaligned sentences. For both WikiSmall and WikiLarge, we reran our experiments removing 5\% and 2\% of the worst alignments.

Finally, we evaluated the models by using the test subsets of external datasets, including: TurkCorpus, ASSET and WikiManual. 

\section{Discussion}

\begin{figure}[t]
\centering
\includegraphics[width=7.5cm]{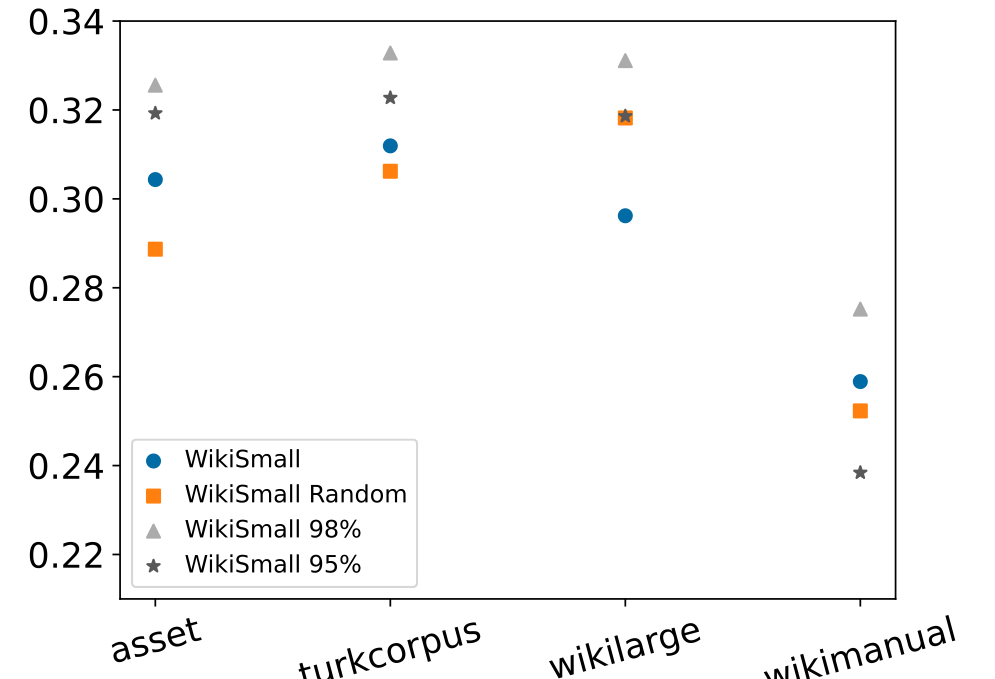}
\caption{SARI scores for evaluating WikiSmall-based models on external test sets.}
\label{fig:sari_wikismall}
\vspace{-3mm} 
\end{figure}

Figure \ref{fig:sari_wikismall} shows the results for WikiSmall. We can see a minor decrease in SARI score with the random splits, which means that the noisy alignments were equivalently present in all the sets rather than using the best cases for training. On the other hand, when the noisy cases are removed from the datasets the increase in model performance is clear.

\begin{figure}[t]
\centering
\includegraphics[width=7.5cm]{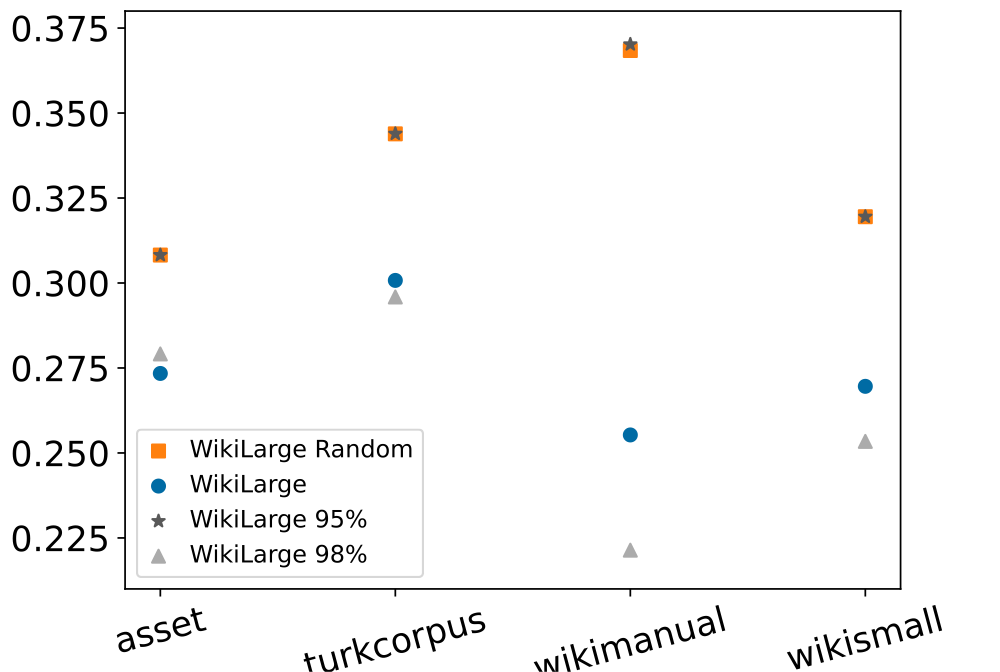}
\caption{SARI scores for evaluating WikiLarge-based models on external test sets.}
\label{fig:sari_wikilarge}
\vspace{-3mm} 
\end{figure}

Likewise, we show WikiLarge results in Figure \ref{fig:sari_wikilarge}. When the data is randomly distributed, we obtain better performance than the original splits. This is consistent with WikiLarge having the largest discrepancy according to our KL-divergence measurements, as shown in Section \ref{sec:kl_divergence_results}. We also found that the 95\% split gave a similar behaviour to WikiLarge Random. Meanwhile, the 98\% dataset, gave a similar performance to the original splits for ASSET and TurkCorpus\footnote{ASSET and Turk Corpus results are an average on their multiple references scores.}. 

We can also note, that although there is a performance difference between WikiSmall Random and WikiSmall 95\%, in WikiLarge the same splits have quite similar results. We believe these discrepancies are related to the size and distribution of the training sets. WikiLarge subset is three times bigger than WikiSmall in the number of simple/complex pairs. Also, WikiLarge has a higher KL-divergence ($\approx$0.46) than WikiSmall ($\approx$0.06), which means that WikiLarge could benefit more from a random distribution experiment than WikiSmall, resulting in higher performance on WikiLarge. Further differences may be caused by the procedures used to make the training/test splits in the original research, which were not described in the accompanying publications. 

Using randomised permutation testing, we have confirmed that the SARI differences between the models based on the original split and our best alternative (95\% refined) is statistically significant ($p < 0.05$) for each configuration discussed above.

In this study, we have shown the limitations of TS datasets and the variations in performance in different splits configurations. In contrast, existing evidence cannot determine which is the most suitable split, especially since this could depend on each specific scenario or target audience (e.g., model data similar to ``real world" applications). 

Also, we have measured our results using SARI, not only because it is the standard evaluation metric in TS but also because there is no better automatic alternatives to measure simplicity. We use SARI as a way to expose and quantify SOTA TS datasets limitations. The increase in SARI scores should be interpreted as the variability in the relative quality of the output simplifications. By relative we mean, that there is a change in simplicity gain but we cannot state the simplification is at its best quality since the metric itself has its own weaknesses.

\section{Conclusions}
\label{conclusions}

In this paper, we have shown 1) the statistical limitations of TS datasets, and 2) the relevance of subset distribution for building more robust models. To our knowledge, distribution-based TS datasets analysis has not been considered before. We hope that the exposure of these limitations kicks off a discussion in the TS community on whether we are in the correct direction regarding evaluation resources in TS and more widely in NLG. The creation of new resources is expensive and complex, however, we have shown that current resources can be refined, motivating future studies in the field of TS.

\section*{Acknowledgments}

We would like to thank Nhung T.H. Nguyen and Jake Vasilakes for their valuable discussions and comments. Laura Vásquez-Rodríguez's work was funded by the \textit{Kilburn Scholarship} from the \textit{University of Manchester}. 
Piotr Przybyła's work was supported by the \textit{Polish National Agency for Academic Exchange} through a \textit{Polish Returns} grant number PPN/PPO/2018/1/00006.

\bibliographystyle{acl_natbib}
\bibliography{anthology,acl2021}

\begin{thebibliography}{32}
\expandafter\ifx\csname natexlab\endcsname\relax\def\natexlab#1{#1}\fi

\bibitem[{Aluisio et~al.(2010)Aluisio, Specia, Gasperin, and
  Scarton}]{Aluisio2010}
Sandra Aluisio, Lucia Specia, Caroline Gasperin, and Carolina Scarton. 2010.
\newblock {Readability assessment for text simplification}.
\newblock \emph{Proceedings of the NAACL HLT 2010 Fifth Workshop on Innovative
  Use of NLP for Building Educational Applications}, pages 1--9.

\bibitem[{Alva-Manchego et~al.(2017)Alva-Manchego, Bingel, Paetzold, Scarton,
  and Specia}]{Alva-Manchego2017}
Fernando Alva-Manchego, Joachim Bingel, Gustavo~H Paetzold, Carolina Scarton,
  and Lucia Specia. 2017.
\newblock \href {https://github.com/jbingel/} {{Learning How to Simplify From
  Explicit Labeling of Complex-Simplified Text Pairs}}.
\newblock \emph{Proceedings of the Eighth International Joint Conference on
  Natural Language Processing (Volume 1: Long Papers)}, pages 295--305.

\bibitem[{Alva-Manchego et~al.(2020{\natexlab{a}})Alva-Manchego, Martin,
  Bordes, Scarton, Sagot, and Specia}]{Alva-Manchego2020b}
Fernando Alva-Manchego, Louis Martin, Antoine Bordes, Carolina Scarton,
  Beno{\^{i}}t Sagot, and Lucia Specia. 2020{\natexlab{a}}.
\newblock \href {https://doi.org/10.18653/v1/2020.acl-main.424} {{ASSET: A
  Dataset for Tuning and Evaluation of Sentence Simplification Models with
  Multiple Rewriting Transformations}}.
\newblock \emph{arXiv}.

\bibitem[{Alva-Manchego et~al.(2020{\natexlab{b}})Alva-Manchego, Martin,
  Bordes, Scarton, Sagot, and Specia}]{alva-manchego-etal-2020-asset}
Fernando Alva-Manchego, Louis Martin, Antoine Bordes, Carolina Scarton,
  Beno{\^\i}t Sagot, and Lucia Specia. 2020{\natexlab{b}}.
\newblock \href {https://doi.org/10.18653/v1/2020.acl-main.424} {{ASSET}: {A}
  dataset for tuning and evaluation of sentence simplification models with
  multiple rewriting transformations}.
\newblock In \emph{Proceedings of the 58th Annual Meeting of the Association
  for Computational Linguistics}, pages 4668--4679, Online. Association for
  Computational Linguistics.

\bibitem[{Caglayan et~al.(2020)Caglayan, Madhyastha, and
  Specia}]{caglayan-etal-2020-curious}
Ozan Caglayan, Pranava Madhyastha, and Lucia Specia. 2020.
\newblock \href {https://www.aclweb.org/anthology/2020.coling-main.210}
  {Curious case of language generation evaluation metrics: A cautionary tale}.
\newblock In \emph{Proceedings of the 28th International Conference on
  Computational Linguistics}, pages 2322--2328, Barcelona, Spain (Online).
  International Committee on Computational Linguistics.

\bibitem[{Cao et~al.(2020)Cao, Shui, Pan, Kan, Liu, and Chua}]{Cao2020}
Yixin Cao, Ruihao Shui, Liangming Pan, Min-Yen Kan, Zhiyuan Liu, and Tat-Seng
  Chua. 2020.
\newblock \href {https://doi.org/10.18653/v1/2020.acl-main.100} {{Expertise
  Style Transfer: A New Task Towards Better Communication between Experts and
  Laymen}}.
\newblock In \emph{arXiv}, pages 1061--1071. Association for Computational
  Linguistics (ACL).

\bibitem[{Cooper and Shardlow(2020)}]{cooper-shardlow-2020-combinmt}
Michael Cooper and Matthew Shardlow. 2020.
\newblock \href {https://www.aclweb.org/anthology/2020.lrec-1.686}
  {{C}ombi{NMT}: An exploration into neural text simplification models}.
\newblock In \emph{Proceedings of the 12th Language Resources and Evaluation
  Conference}, pages 5588--5594, Marseille, France. European Language Resources
  Association.

\bibitem[{{De Belder} and Moens(2010)}]{DeBelder2010}
Jan {De Belder} and Marie-Francine Moens. 2010.
\newblock \href
  {http://scholar.google.com/scholar?hl=en{\&}btnG=Search{\&}q=intitle:Text+Simplification+for+Children{\#}0}
  {{Text Simplification for Children}}.
\newblock \emph{Proceedings of the SIGIR Workshop on Accessible Search
  Systems}, pages 19--26.

\bibitem[{Dong et~al.(2020)Dong, Li, Rezagholizadeh, and Cheung}]{Dong2020}
Yue Dong, Zichao Li, Mehdi Rezagholizadeh, and Jackie Chi~Kit Cheung. 2020.
\newblock \href {http://arxiv.org/abs/1906.08104} {{Editnts: An neural
  programmer-interpreter model for sentence simplification through explicit
  editing}}.
\newblock In \emph{ACL 2019 - 57th Annual Meeting of the Association for
  Computational Linguistics, Proceedings of the Conference}, pages 3393--3402.
  Association for Computational Linguistics (ACL).

\bibitem[{Guo et~al.(2018)Guo, Pasunuru, and Bansal}]{Guo2018}
Han Guo, Ramakanth Pasunuru, and Mohit Bansal. 2018.
\newblock \href {http://arxiv.org/abs/1806.07304} {{Dynamic Multi-Level
  Multi-Task Learning for Sentence Simplification}}.
\newblock In \emph{Proceedings of the 27th International Conference on
  Computational Linguistics (COLING 2018)}, pages 462--476.

\bibitem[{Hou et~al.(2019)Hou, Jochim, Gleize, Bonin, and
  Ganguly}]{hou-etal-2019-identification}
Yufang Hou, Charles Jochim, Martin Gleize, Francesca Bonin, and Debasis
  Ganguly. 2019.
\newblock \href {https://doi.org/10.18653/v1/P19-1513} {Identification of
  tasks, datasets, evaluation metrics, and numeric scores for scientific
  leaderboards construction}.
\newblock In \emph{Proceedings of the 57th Annual Meeting of the Association
  for Computational Linguistics}, pages 5203--5213, Florence, Italy.
  Association for Computational Linguistics.

\bibitem[{Hwang et~al.(2015)Hwang, Hajishirzi, Ostendorf, and Wu}]{Hwang2015}
William Hwang, Hannaneh Hajishirzi, Mari Ostendorf, and Wei Wu. 2015.
\newblock \href {https://doi.org/10.3115/v1/n15-1022} {{Aligning sentences from
  standard Wikipedia to simple Wikipedia}}.
\newblock In \emph{NAACL HLT 2015 - 2015 Conference of the North American
  Chapter of the Association for Computational Linguistics: Human Language
  Technologies, Proceedings of the Conference}, pages 211--217. Association for
  Computational Linguistics (ACL).

\bibitem[{Inui et~al.(2003)Inui, Fujita, Takahashi, Iida, and
  Iwakura}]{Inui2003}
Kentaro Inui, Atsushi Fujita, Tetsuro Takahashi, Ryu Iida, and Tomoya Iwakura.
  2003.
\newblock \href {https://doi.org/10.3115/1118984.1118986} {{Text Simplification
  for Reading Assistance: A Project Note}}.
\newblock In \emph{Proceedings of the Second International Workshop on
  Paraphrasing - Volume 16}, PARAPHRASE '03, pages 9--16, USA. Association for
  Computational Linguistics (ACL).

\bibitem[{Jiang et~al.(2020)Jiang, Maddela, Lan, Zhong, and Xu}]{Jiang2020}
Chao Jiang, Mounica Maddela, Wuwei Lan, Yang Zhong, and Wei Xu. 2020.
\newblock \href {https://doi.org/10.18653/v1/2020.acl-main.709} {{Neural CRF
  Model for Sentence Alignment in Text Simplification}}.
\newblock In \emph{arXiv}, pages 7943--7960. arXiv.

\bibitem[{Kullback and Leibler(1951)}]{Kullback1951}
S.~Kullback and R.~A. Leibler. 1951.
\newblock \href {https://doi.org/10.1214/aoms/1177729694} {{On Information and
  Sufficiency}}.
\newblock \emph{The Annals of Mathematical Statistics}, 22(1):79--86.

\bibitem[{van~der Lee et~al.(2019)van~der Lee, Gatt, van Miltenburg, Wubben,
  and Krahmer}]{van-der-lee-etal-2019-best}
Chris van~der Lee, Albert Gatt, Emiel van Miltenburg, Sander Wubben, and Emiel
  Krahmer. 2019.
\newblock \href {https://doi.org/10.18653/v1/W19-8643} {Best practices for the
  human evaluation of automatically generated text}.
\newblock In \emph{Proceedings of the 12th International Conference on Natural
  Language Generation}, pages 355--368, Tokyo, Japan. Association for
  Computational Linguistics.

\bibitem[{Morgan(2006)}]{Morgan2006}
William Morgan. 2006.
\newblock \href {https://cs.stanford.edu/people/wmorgan/sigtest.pdf}
  {{Statistical Hypothesis Tests for NLP or: Approximate Randomization for Fun
  and Profit}}.

\bibitem[{Nisioi et~al.(2017)Nisioi, {\v{S}}tajner, Ponzetto, and
  Dinu}]{Nisioi2017}
Sergiu Nisioi, Sanja {\v{S}}tajner, Simone~Paolo Ponzetto, and Liviu~P. Dinu.
  2017.
\newblock \href {https://doi.org/10.18653/v1/P17-2014} {{Exploring neural text
  simplification models}}.
\newblock In \emph{ACL 2017 - 55th Annual Meeting of the Association for
  Computational Linguistics, Proceedings of the Conference (Long Papers)},
  volume~2, pages 85--91. Association for Computational Linguistics (ACL).

\bibitem[{Pang(2019)}]{Pang2019}
Richard~Yuanzhe Pang. 2019.
\newblock \href {http://arxiv.org/abs/1910.03747} {{The Daunting Task of
  Real-World Textual Style Transfer Auto-Evaluation}}.
\newblock \emph{arXiv}.

\bibitem[{Papineni et~al.(2001)Papineni, Roukos, Ward, and Zhu}]{Papineni2002}
Kishore Papineni, Salim Roukos, Todd Ward, and Wei-Jing Zhu. 2001.
\newblock \href {https://doi.org/10.3115/1073083.1073135} {{BLEU: a method for
  automatic evaluation of machine translation}}.
\newblock \emph{ACL}, Proceedings of the 40th Annual Meeting of the Association
  for Computational Linguistics(July):311--318.

\bibitem[{Rello et~al.(2013)Rello, Baeza-Yates, Bott, and Saggion}]{Rello2013}
Luz Rello, Ricardo Baeza-Yates, Stefan Bott, and Horacio Saggion. 2013.
\newblock \href {https://doi.org/10.1145/2461121.2461126} {{Simplify or help?
  Text simplification strategies for people with dyslexia}}.
\newblock In \emph{W4A 2013 - International Cross-Disciplinary Conference on
  Web Accessibility}.

\bibitem[{Scarton and Specia(2018)}]{Scarton2018}
Carolina Scarton and Lucia Specia. 2018.
\newblock \href {https://doi.org/10.18653/v1/p18-2113} {{Learning
  simplifications for specific target audiences}}.
\newblock In \emph{ACL 2018 - 56th Annual Meeting of the Association for
  Computational Linguistics, Proceedings of the Conference (Long Papers)},
  volume~2, pages 712--718, Stroudsburg, PA, USA. Association for Computational
  Linguistics.

\bibitem[{Shardlow(2014)}]{Shardlow2014}
Matthew Shardlow. 2014.
\newblock \href {https://doi.org/10.14569/specialissue.2014.040109} {{A Survey
  of Automated Text Simplification}}.
\newblock \emph{International Journal of Advanced Computer Science and
  Applications}, 4(1).

\bibitem[{Silveira and Branco(2012)}]{Silveira2012}
Sara~Botelho Silveira and Ant{\'{o}}nio Branco. 2012.
\newblock {Enhancing multi-document summaries with sentence simplification}.
\newblock In \emph{Proceedings of the 2012 International Conference on
  Artificial Intelligence, ICAI 2012}, volume~2, pages 742--748.

\bibitem[{Sulem et~al.(2018)Sulem, Abend, and Rappoport}]{Sulem2018}
Elior Sulem, Omri Abend, and Ari Rappoport. 2018.
\newblock \href {https://doi.org/10.18653/v1/D18-1081} {{BLEU is Not Suitable
  for the Evaluation of Text Simplification}}.
\newblock In \emph{Proceedings of the 2018 Conference on Empirical Methods in
  Natural Language Processing}, pages 738--744, Stroudsburg, PA, USA.
  Association for Computational Linguistics.

\bibitem[{Surya et~al.(2019)Surya, Mishra, Laha, Jain, and
  Sankaranarayanan}]{Surya2019}
Sai Surya, Abhijit Mishra, Anirban Laha, Parag Jain, and Karthik
  Sankaranarayanan. 2019.
\newblock \href {http://arxiv.org/abs/1810.07931} {{Unsupervised Neural Text
  Simplification}}.
\newblock \emph{ACL 2019 - 57th Annual Meeting of the Association for
  Computational Linguistics, Proceedings of the Conference}, pages 2058--2068.

\bibitem[{Vu et~al.(2018)Vu, Hu, Munkhdalai, and Yu}]{Vu2018}
Tu~Vu, Baotian Hu, Tsendsuren Munkhdalai, and Hong Yu. 2018.
\newblock \href {https://doi.org/10.18653/v1/n18-2013} {{Sentence
  simplification with memory-augmented neural networks}}.
\newblock In \emph{NAACL HLT 2018 - 2018 Conference of the North American
  Chapter of the Association for Computational Linguistics: Human Language
  Technologies - Proceedings of the Conference}, volume~2, pages 79--85.
  Association for Computational Linguistics (ACL).

\bibitem[{Wagner and Fischer(1974)}]{Wagner1974}
Robert~A. Wagner and Michael~J. Fischer. 1974.
\newblock \href {https://doi.org/10.1145/321796.321811} {{The String-to-String
  Correction Problem}}.
\newblock \emph{Journal of the ACM (JACM)}, 21(1):168--173.

\bibitem[{Xu et~al.(2015)Xu, Callison-Burch, and Napoles}]{Xu2015}
Wei Xu, Chris Callison-Burch, and Courtney Napoles. 2015.
\newblock \href {https://doi.org/10.1162/tacl_a_00139} {{Problems in Current
  Text Simplification Research: New Data Can Help}}.
\newblock \emph{Transactions of the Association for Computational Linguistics},
  3:283--297.

\bibitem[{Xu et~al.(2016)Xu, Napoles, Pavlick, Chen, and
  Callison-Burch}]{Xu2016}
Wei Xu, Courtney Napoles, Ellie Pavlick, Quanze Chen, and Chris Callison-Burch.
  2016.
\newblock \href {https://doi.org/10.1162/tacl_a_00107} {{Optimizing Statistical
  Machine Translation for Text Simplification}}.
\newblock \emph{Transactions of the Association for Computational Linguistics},
  4:401--415.

\bibitem[{Zhang and Lapata(2017)}]{Zhang2017}
Xingxing Zhang and Mirella Lapata. 2017.
\newblock \href {https://doi.org/10.18653/v1/d17-1062} {{Sentence
  Simplification with Deep Reinforcement Learning}}.
\newblock In \emph{EMNLP 2017 - Conference on Empirical Methods in Natural
  Language Processing, Proceedings}, pages 584--594. Association for
  Computational Linguistics (ACL).

\bibitem[{Zhao et~al.(2018)Zhao, Meng, He, Andi, and Bambang}]{Zhao2018}
Sanqiang Zhao, Rui Meng, Daqing He, Saptono Andi, and Parmanto Bambang. 2018.
\newblock \href {https://doi.org/10.18653/v1/d18-1355} {{Integrating
  transformer and paraphrase rules for sentence simplification}}.
\newblock In \emph{Proceedings of the 2018 Conference on Empirical Methods in
  Natural Language Processing, EMNLP 2018}, pages 3164--3173. Association for
  Computational Linguistics.

\end{thebibliography}

\end{document}